\documentclass{bmvc2k}

\usepackage[utf8]{inputenc} 
\usepackage[T1]{fontenc}    
\usepackage{hyperref}       
\usepackage{url}            
\usepackage{booktabs}       
\usepackage{amsfonts}       
\usepackage{nicefrac}       
\usepackage{microtype}      
\usepackage[table]{xcolor}
\usepackage{colortbl}
\usepackage{amsthm} 
\usepackage{multirow}
\usepackage{makecell}
\usepackage{array}
\usepackage{graphicx}
\usepackage{amsmath}

\usepackage{wrapfig}
\usepackage{algorithm}
\usepackage{algpseudocode}
\usepackage{microtype}
\usepackage{booktabs}
\usepackage{wrapfig}

\newcommand{\best}[1]{\cellcolor{red!25}\textbf{#1}}
\newcommand{\second}[1]{\cellcolor{yellow!25}\underline{#1}}
\makeatletter
\colorlet{bmvcaptionblue}{bmv@captioncolor}
\makeatother

\title{CG-GLORE: A Conjugate Gradient–Based Global-Local Regularization Network for Sparse-View CT Reconstruction}
\addauthor{Tran Xuan Hieu Le}{le.tran_xuan_hieu.lx0@naist.ac.jp}{1}
\addauthor{Doanh C. Bui}{doanhbc@uit.edu.vn}{2,3}
\addauthor{Vu Trung Duong Le}{le.duong@naist.ac.jp}{1,2,3}
\addauthor{Hoai Luan Pham}{pham.luan@is.naist.jp}{1,2,3}
\addauthor{Khang Nguyen}{khangnttm@uit.edu.vn}{2,3}
\addauthor{Ma\"{i} K. Nguyen}{mai.nguyen-verger@cyu.fr}{4}
\addauthor{Tu Bao Ho}{bao@viasm.edu.vn}{5}
\addauthor{Yasuhiko Nakashima}{nakashim@is.naist.jp}{1}

\addinstitution{
 Nara Institute of Science \\and Technology \\
 Nara, Japan
}
\addinstitution{
 University of Information Technology \\
 Ho Chi Minh City, Vietnam 
}
\addinstitution{
 Vietnam National University \\
 Ho Chi Minh City, Vietnam 
}
\addinstitution{
 ETIS, CY Cergy Paris University, ENSEA, CNRS, UMR 8051\\ 
 Cergy-Pontoise, France
}
\addinstitution{
 Vietnam Institute for Advanced Study \\ in Mathematics \\ 
 Hanoi, Vietnam
}

\runninghead{Le et al.}{CG-GLORE for Sparse-View CT Reconstruction}

\begin{document}

\maketitle

\begin{abstract}
Sparse-view computed tomography (CT) reduces radiation dose by acquiring fewer projection views, but the resulting inverse problem is highly ill-posed and often produces severe streak artifacts. Existing deep reconstruction methods have achieved promising performance, yet many rely on first-order updates or large regularization networks, which can be less effective in ill-conditioned settings. We propose \textbf{CG-GLORE}, a compact deep unrolling framework inspired by second-order optimization for sparse-view CT reconstruction. Each unrolled stage uses a CG-solved linear system based on a structured Hessian surrogate: it retains the physics-induced curvature of the data-fidelity term while using an identity approximation for the learned regularization term. Thus, the method is second-order-inspired rather than an exact Newton method for the full learned objective. To model image priors, we design a Global-Local Regularization Network (GLORE), which combines convolutional local feature extraction with a Long-Range Dependency Representation module based on sparse patchification and Nystr\"{o}m attention. This design captures anatomical details and non-local dependencies while maintaining practical complexity. Experiments on AAPM and DeepLesion under multiple sparse-view and noise settings show that CG-GLORE achieves strong quantitative performance, stable convergence, lower noise power, and improved visual fidelity compared with representative reconstruction methods. 
\end{abstract}

\section{Introduction}
\label{sec:intro}
Computed tomography (CT) is essential for clinical diagnosis, but repeated exposure to ionizing radiation raises safety concerns \cite{wang2008outlook}. Sparse-view CT reduces the number of projection views to lower radiation dose \cite{katsura2012model}, yet the resulting incomplete measurements make reconstruction highly ill-posed and often introduce severe streak artifacts, especially when using Filtered Back Projection (FBP) \cite{cormack1963representation}.

Deep learning has improved sparse-view CT reconstruction by post-processing FBP images or jointly exploiting image and sinogram domains \cite{chao2022sparse,wang2024low,tao2021learning,wu2021noise}. However, many of these methods remain weakly coupled with the reconstruction physics and rely heavily on local image statistics. Deep unrolling networks address this by embedding iterative optimization into learnable architectures, but most existing designs still use first-order updates, which can be inefficient for ill-conditioned sparse-view reconstruction and often require large regularization networks to compensate for weak optimization directions.

We propose \textbf{CG-GLORE}, a compact second-order-inspired unrolling framework for sparse-view CT reconstruction. Each stage uses a structured Hessian surrogate that combines the analytic data-fidelity curvature with an identity approximation for the learned regularization term. Instead of computing or storing an exact Hessian, we solve the resulting linear system by Conjugate Gradient (CG). This formulation is intended as a stable, physics-aware surrogate update, rather than as an exact Newton step for the full learned objective.

To model the learned regularization term, we further design the Global-Local Regularization Network (GLORE). GLORE combines convolutional layers for local anatomical details with a Long-Range Dependency Representation (LORAD) module for non-local structure modeling. LORAD uses sparse patchification and Nystr\"{o}m-based approximate self-attention to capture long-range dependencies efficiently. Together, CG and GLORE separate the roles of optimization and prior modeling: CG provides effective update directions, while GLORE focuses on image regularity.

Our contributions are summarized as follows:
\begin{itemize}
    \item We introduce \textbf{CG-GLORE}, a second-order-inspired deep unrolling framework that formulates each reconstruction stage as a CG-solvable linear system based on a structured Hessian surrogate.
    \item We design \textbf{GLORE}, a compact global-local regularization network that combines convolutions with efficient Nystr\"{o}m-based long-range dependency modeling.
    \item We validate CG-GLORE under different sparse-view and noise settings, showing improved reconstruction quality and favorable optimization behavior.
\end{itemize}
\section{Related Work}
\noindent\textbf{Deep learning for sparse-view CT reconstruction.}
Early deep learning-based methods for sparse-view CT reconstruction commonly treat the problem as an image restoration task, where an FBP reconstruction is enhanced by a neural network. Representative approaches employ convolutional architectures to suppress streak artifacts and recover anatomical structures in the image domain \cite{jin2017deep,tao2021learning,wu2021noise}. To further exploit the measurement process, dual-domain methods extend this idea by incorporating sinogram-domain information, allowing networks to interpolate or refine projection data before or during image reconstruction \cite{chao2022sparse,wu2025dual}. Although these methods achieve promising visual quality, they are often weakly coupled with the underlying optimization model and may struggle to enforce data consistency under highly sparse or noisy acquisition settings.

\noindent\textbf{Deep unrolling and learned regularization.}
Deep unrolling networks explicitly embed iterative optimization into neural architectures by unfolding reconstruction updates into a finite number of learnable stages. Learned Primal-Dual \cite{adler2018learned} and LEARN \cite{chen2018learn} demonstrate the effectiveness of learning update rules and regularization terms for CT reconstruction. More recent methods introduce non-local modeling into the regularization module, such as Transformer-based local--nonlocal regularization in RegFormer \cite{xia2023regformer} or quasi-Newton optimization with an MLP-Mixer-style regularizer in QN-Mixer \cite{ayad2024qn}. These studies show the importance of both global context modeling and improved optimization dynamics. In contrast, CG-GLORE uses a CG-solvable structured surrogate that preserves the analytic data-fidelity curvature, without estimating the curvature of the learned regularizer, while GLORE models local structures and long-range dependencies efficiently.

\section{Methodology}
\label{sec:method}

\subsection{Preliminary}
\label{subsec:preli}

\noindent\textbf{Inverse Problem for CT Reconstruction.}
In computed tomography (CT), the reconstruction task can be formulated as an \textit{inverse problem}. 
Given a sinogram $\mathbf{y} \in \mathbb{R}^{m}$, where $m = n_v \times n_d$, with $n_v$ denoting the number of projection views and $n_d$ the number of detector elements, the CT image $\mathbf{x} \in \mathbb{R}^{n}$, with $n = H \times W$, where $H$ and $W$ denote the height and width of the image, respectively, is to be reconstructed. This image characterizes the internal structure of the scanned object. The sinogram is obtained via the forward projection process:
\begin{equation}
\mathbf{y} = \mathbf{A}\mathbf{x} + {\epsilon},
\label{eq:forward}
\end{equation}
where $\mathbf{A}  \in \mathbb{R}^{m \times n}$ denotes the Radon transform operator (i.e., the CT system matrix) and ${\epsilon}$ represents measurement noise. The reconstructed image $\mathbf{x}$ is typically expressed in Hounsfield Units (HU). Eq. \ref{eq:forward} is the forward model; recovering $\mathbf{x}$ from incomplete $\mathbf{y}$ is the corresponding ill-posed inverse problem. To obtain the optimal reconstruction $\hat{\mathbf{x}}$, an iterative optimization procedure is employed to minimize the following objective function:

\begin{equation}
\hat{\mathbf{x}} = \underset{\mathbf{x}}{\arg\min} \text{ } J(\mathbf{x}) = \frac{\lambda}{2}\Vert \mathbf{A}\mathbf{x}-\mathbf{y}\rVert^2_2+\mathcal{R}(\mathbf{x}), \nonumber
\end{equation}

\noindent where $\mathcal{R}(\mathbf{x})$ is a regularization term that constraint the image with the expected conditions to obtain better $\hat{\mathbf{x}}$. For each iteration, the reconstructed image is updated thanks to simulated gradient descent:

\begin{equation}
\mathbf{x}_{t+1} = \mathbf{x}_t - \alpha\underbrace{\left[\lambda \mathbf{A}^{T}(\mathbf{A}\mathbf{x}_t-\mathbf{y}) + \nabla_{\mathbf{x}}\mathcal{R}(\mathbf{x}_t)\right]}_{\nabla_{\mathbf{x}}J(\mathbf{x}_t)}, \quad t = 1, \dots, T
\label{eq:gradient}
\end{equation}
where $T$ denotes the number of iterations. There are many forms of the regularization term $\mathcal{R}(\mathbf{x}_t)$ \cite{jin2017deep,wang2022dudotrans,su2022generalized,wu2025dual} appear in Eq.~\ref{eq:gradient}.

\noindent\textbf{Sparse-view CT reconstruction.} This aims to reconstruct a high-quality CT image $\hat{\mathbf{x}}$ from sparse projection views $\mathbf{y}' \in \mathbb{R}^{n'_v \times n_d}$, where $n'_v < n_v$. Typically, deep unrolling networks leverage deep learning to construct a learnable module $\mathcal{G}_\theta(\cdot)$ to approximate the gradient term $\nabla_{\mathbf{x}} J(\mathbf{x}_t)$. The parameters $\theta$ are optimized by minimizing a loss function that measures the discrepancy between the ground-truth image reconstructed from full-view projections $\mathbf{x}_{gt}$ and the reconstructed image $\hat{\mathbf{x}}$ obtained from sparse views.

\subsection{Second-Order-Inspired Update}

Most deep unrolling methods update the reconstruction using a first-order gradient descent step \cite{adler2018learned,chen2018learn,su2022generalized,xia2023regformer,wu2025dual}. Although this update rule is simple and efficient, first-order optimization may converge slowly when solving ill-conditioned inverse problems such as sparse-view CT reconstruction.
To improve the update direction, we use an iterative mechanism inspired by Newton-type optimization. The following expression provides a conceptual Newton-type form; in our method, the exact Hessian is replaced by the structured surrogate described in Sec.~\ref{subsec:conjug}. Thus, we do not claim an exact Newton update for the full learned objective.
\begin{equation}
\mathbf{x}_{t+1} = \mathbf{x}_{t} - \alpha_t \mathbf{H}_t^{-1}\nabla_{\mathbf{x}} J(\mathbf{x}_t),
\label{eq:newton}
\end{equation}

\noindent where $\mathbf{H}_t \in \mathbb{R}^{n \times n}$ denotes the structured Hessian surrogate used at iteration $t$. 
The parameter $\alpha_t$ represents the search step and is commonly used for damping or globalization in Newton-type methods \cite{galantai2000theory,nocedal2006numerical}. 
In this work, we set $\alpha_t=1$ for the surrogate update and solve the resulting linear system with CG, avoiding the explicit computation of $\mathbf{H}_t^{-1}$ \cite{hestenes1952cg,nocedal2006numerical,royer2020newtoncg}.

\subsection{Alternative Global-Local Regularization Network}
In this section, we present the proposed Alternative Global-Local (GLORE) Regularization Network for $\mathcal{G}_\theta(\mathbf{x}_t) \approx \nabla_{\mathbf{x}} \mathcal{R}(\mathbf{x}_t)$. GLORE includes two components: 1) \textit{Learnable Convolution Layers} to exploit local patterns in the reconstructed image at the $i^{th}$ iteration ($x_i$) and 2) \textit{Long-Range Dependency Representation \textcolor{black}{(LORAD)}} to explore relationships between distant parts within $x_i$ and to enhance a robust representation for the regularization term.

\subsubsection{Overall Architecture of GLORE} GLORE includes three convolutional layers, with \textcolor{black}{LORAD} placed between the first convolutional layer and the subsequent two convolutional layers ($conv$). In this manner, given an image $x_i \in \mathbb{R}^{H \times W \times 3}$, where $H$ and $W$ are height and width, respectively ($H = W$), the first convolutional layer is learned to exploit local patterns and produce feature maps $F_1 \in \mathbb{R}^{H \times W \times C_1}$, where $C_1$ is the dimensionality of $conv_1$. $F_1$ is then patchified as a sequence of tokens and processed by \textcolor{black}{LORAD} to explore long-range dependencies and capture relations between distant parts, thereby facilitating global patterns. Finally, two more convolutional layers are adopted to further produce high-level feature maps. 

\begin{figure}[h]
    \centering
    \includegraphics[width=1\textwidth]{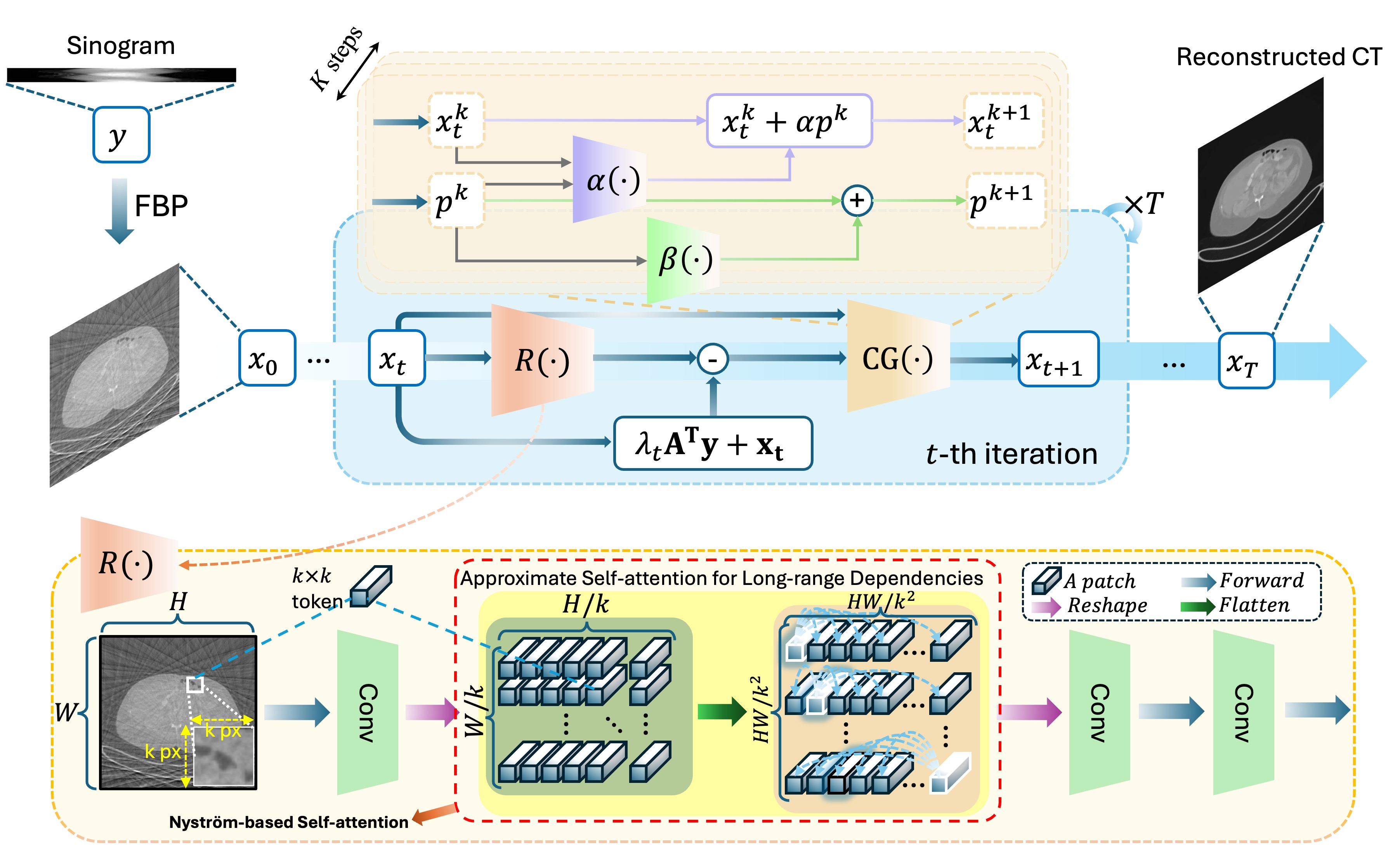}
    \vspace{-2em}
    \caption{\textbf{Overview of the proposed CG-GLORE framework.} Starting from an FBP reconstruction $x_0$, CG-GLORE performs $T$ unrolled reconstruction iterations. At each iteration, a Conjugate Gradient solver computes an update from the structured Hessian surrogate, while the global-local regularization network captures local structures and long-range dependencies to further refine the reconstruction.}
    \label{fig:overview}
\end{figure}

\subsubsection{Long-Range Dependency Representation} 

\noindent\textbf{Sparse Patchification.} Given the feature map $F_1$ produced by the initial convolutional layer $conv_1$, a sparse patchification transforms it into a sequence of tokens $z \in \mathbb{R}^{N \times C_1'}$, where $N$ is the number of tokens and $C_1'$ is the feature dimension of each token. As observed in \cite{nguyen2024image}, increasing the number of tokens generally improves performance. Therefore, we aim to generate as many tokens as possible while maintaining computational efficiency. However, using $N = HW$ tokens results in a quadratic complexity of $\mathcal{O}(N^2)$ when computing pairwise attention, which is computationally prohibitive. To address this, we introduce a \textit{dilated patchification} strategy. Specifically, we apply a sliding window of size $k \times k$ with a \textbf{dilation factor} of $H / k$ (or $W / k$) and stride $s=1$ to group distant spatial values while ensuring that each patch remains non-overlapping. Then, the tokens within each window are flattened into $z_i \in \mathbb{R}^{C_1 \times k^2}$. For convenience, we denote $C_1' = C_1 \times k^2$, effectively compressing the spatial information into the channel dimension. This transformation reduces $N$ to $N/k^2$, leading to a decrease in the pairwise attention complexity from $\mathcal{O}(N^2)$ to $\mathcal{O}(N^2 / k^4)$, significantly improving computational efficiency. We further define $\tilde{N} = N / k^2$ to denote the reduced number of tokens. This approach facilitates the capture of long-range dependencies while preserving essential structural information. The sparse patchification process is detailed in the PyTorch-like Algorithm \ref{algo:1}.

\begin{algorithm}
\caption{Sparse Patchification}
\label{algo:1}
\begin{algorithmic}[1]
\Require $x \in \mathbb{R}^{B \times C_1 \times H \times W}$, input image tensor
\Ensure Patchified representation $z \in \mathbb{R}^{(N/k)\times(C\times k^2)}$

\State \textbf{Function} Patchify($x$)
\Comment{Convert image into patch sequence}

\State $(B, C_1, H, W) \gets x.\texttt{shape}$
\Comment{$B$: Batch size, $C$: Channels, $H \times W$: Image size}

\State $x \gets x.\texttt{unfold}(2, H/k, W/k).\texttt{unfold}(3, H/k, W/k)$
\Comment{Extract patches of size $k \times k$ with dilation factor of $H/k$}

\State $x \gets x.\texttt{view}(B, -1, (H/k)\times(W/k))$
\Comment{Reshape to sequence of patches}

\State \textbf{return} $x.\texttt{permute}(0, 2, 1)$
\Comment{Rearrange to (batch, patches, channels)}

\end{algorithmic}
\end{algorithm}

\noindent\textbf{Long-Range Dependency Computation.} For modeling long-range dependencies, self-attention \cite{vaswani2017attention} has demonstrated strong performance. The attention map is computed as $\mathcal{A} = \text{Softmax}\big(\frac{QK^\top}{\sqrt{C'_1}}\big)$, where $Q = W^Q z$, $K = W^K z$, and $V = W^V z$. Here, $W^Q \in \mathbb{R}^{C'_1 \times d_q}$, $W^K \in \mathbb{R}^{C'_1 \times d_k}$, and $W^V \in \mathbb{R}^{C'_1 \times d_v}$ are learnable projection matrices, while $d_q$, $d_k$, and $d_v$ denote the dimensions of queries $Q$, keys $K$, and values $V$, respectively.

\noindent However, computing the attention matrix $\mathcal{A}$ remains computationally intensive, \textit{i.e.}, $\mathcal{O}(\tilde{N})$. To address this, we adopt an approximate self-attention approach that estimates pairwise relational similarities between tokens while preserving as much feature information as possible. We follow \cite{wang2013improving,xiong2021nystromformer}, which demonstrates that $\mathcal{A}$ can be approximated by selecting $L$ landmarks from $Q$ and $K$ based on Nystr\"{o}m method. Following \cite{shen2018baseline}, landmarks are selected by segmenting tokens in $Q$ or $K$ in a left-to-right order.  
To ensure even segmentation, the token sequence $z$ is first padded with zero values:

\begin{equation}
z' = [z \Vert \mathbf{0}_{M}], \quad \text{where } M = \left\lceil \frac{\tilde{N}}{L} \right\rceil L - \tilde{N},
\end{equation}

\noindent $[\cdot || \cdot]$ is the concatenation operation. Next, three learnable projection matrices, $W^Q$, $W^K$, and $W^V$, are applied to $z'$ to obtain the queries $Q$, keys $K$, and values $V$. The tokens are then segmented, where $i^{th}$ segment is defined as follows:

\begin{equation}
S^Q_i = \{q_j\}_{j=s_i}^{e_i}, \quad S^K_i = \{k_j\}_{j=s_i}^{e_i},
\end{equation}

\noindent where $S^Q_i$ and $S^K_i$ denote $i^{th}$ segment in $Q$ and $K$, respectively. $s_i$ and $e_i$ denote the start and end indices of the tokens $q_j\in Q$ or $k_j \in K$. These indices are defined as $s_i = \sum_{k=1}^{i-1} \vert S^Q_k \vert + 1$ and $e_i = \sum_{k=1}^{i} \vert S^Q_k \vert$, where $\vert S^Q_k \vert$ denotes the number of tokens in $k^{th}$ segment. Finally, each landmark is formed by taking the mean of all tokens in the corresponding segment. This creates a set of landmarks for $Q$ and $K$, denoted as $\tilde{Q}$ and $\tilde{K}$, respectively:

\begin{equation}
\tilde{Q} = \left\{ \frac{1}{\vert S^Q_i \vert} \sum_{q \in S^Q_i} q \right\}_{i=1}^{L}, \quad \tilde{K} = \left\{ \frac{1}{\vert S^K_i \vert} \sum_{k \in S^K_i} k \right\}_{i=1}^{L}.
\end{equation}

\noindent Then, the computation of $\mathcal{A}$ is decomposed and approximated in low-rank manner as follows:

\begin{equation}
\tilde{\mathcal{A}} = Q\tilde{K}^{\top} \cdot (\tilde{Q}\tilde{K}^{\top})^{+}  \cdot \tilde{Q}{K}^{\top}.
\label{eq:17}
\end{equation}

\noindent where $(\cdot)^{+}$ denotes the Moore-Penrose pseudoinverse \cite{razavi2014new}. As the number of landmarks $L \ll \tilde{N}$, Eq. \ref{eq:17} reduces the computational complexity of the attention map from $\mathcal{O}(\tilde{N}^2)$ to $\mathcal{O}(\tilde{N})$, which further facilitates long-range dependency across as many tokens as possible while maintaining feasible complexity. We then obtain the long-range dependency representation by multiplying $z'$ with $\tilde{\mathcal{A}}$: $z'' = V \odot \tilde{\mathcal{A}}$.

\noindent We also apply multi-head self-attention, where $Q$, $K$, and $V$ are decomposed into $h$ heads. This means that we apply Eq. \ref{eq:17} $h$ times and then concatenate the results. $z'$ then undergoes another linear projection for alignment. Notably, we define $d_q = d_k = d_v = C_1'$. 

\noindent\textbf{Patch Extension \& Skip Connection.} After obtaining $z''$, we reshape each token $z''_i \in \mathbb{R}^{C_1'}$ into tensor form $F_{1,i}' \in \mathbb{R}^{k \times k \times C_1}$. In this manner, we convert $z'' \in \mathbb{R}^{\tilde{N} \times C_1'}$ into $F_1' \in \mathbb{R}^{H \times W \times C_1}$. This process is presented as PyTorch-like pseudocode in Algorithm \ref{algo:2}. A skip connection is performed on $F_1'$ and $F_1$, after which the result undergoes the remaining convolution layers with ReLU activation in between:

\begin{equation}
F_{final} = conv_3\big(ReLU(conv_2(F_1+F_1'))\big).
\end{equation}

\begin{algorithm}
\caption{Patch Extension}
\label{algo:2}
\begin{algorithmic}[1]
\Require Patchified representation $z'' \in \mathbb{R}^{(N/k)\times(C\times k^2)}$
\Ensure $x \in \mathbb{R}^{B \times C_1 \times H \times W}$

\State \textbf{Function} PatchExtend($z''$)
\Comment{Reconstruct image from patch sequence}

\State $B \gets z''.\texttt{shape}[0]$
\Comment{Batch size}

\State $z'' \gets z''.\texttt{permute}(0,2,1).\texttt{view}(B, C_1', H/k, W/k)$
\Comment{Rearrange patches to 2D}

\State $z'' \gets z''.\texttt{view}(B, C, k, k, H/k, W/k)$
\Comment{Reshape to original image space}

\State \textbf{return} $z''.\texttt{permute}(0,1,2,4,3,5).\texttt{view}(B, C_1, H, W)$
\Comment{Restore image format}

\end{algorithmic}
\end{algorithm}

\subsubsection{Hessian Approximation.}

Explicitly computing the Hessian $\nabla_x^2 J(\mathbf{x}_t) \in \mathbb{R}^{n \times n}$ is computationally infeasible in large-scale CT reconstruction due to the high dimensionality ($n = H \times W$). To obtain a practical update, we use a structured Hessian surrogate that retains the analytic, physics-induced curvature of the data-fidelity term while treating the learned module $\mathcal{G}_\theta(\cdot)$ as a gradient correction.

\noindent\textbf{Local Linear Approximation.}
The learned gradient representation $\mathcal{G}_\theta(\cdot)$ is locally approximated via first-order expansion:
\begin{equation}
\mathcal{G}_\theta(\mathbf{x}_t + \Delta \mathbf{x})
\approx 
\mathcal{G}_\theta(\mathbf{x}_t) 
+ \nabla_{\mathbf{x}} \mathcal{G}_\theta(\mathbf{x}_t)\,\Delta \mathbf{x},
\end{equation}
where $\nabla_{\mathbf{x}} \mathcal{G}_\theta(\mathbf{x}_t)$ denotes the Jacobian matrix. While this Jacobian would contribute to the curvature of the full learned objective, it is not explicitly evaluated in our method.

\noindent\textbf{Identity Surrogate Approximation.}
Motivated by prior works in residual learning and damped second-order optimization, where identity mappings or diagonal regularization are employed to stabilize optimization and improve conditioning~\cite{he2016identity,levenberg1944method,marquardt1963algorithm}, 
we use an identity surrogate for the local action of the Jacobian:
\begin{equation}
\nabla_{\mathbf{x}} \mathcal{G}_\theta(\mathbf{x}_t)\,\Delta \mathbf{x} \approx \Delta \mathbf{x}.
\end{equation}
This approximation treats $\mathcal{G}_\theta(\cdot)$ as a stable residual correction and avoids estimating learned-prior curvature.

\noindent\textbf{Structured Hessian Form.}
The following formal Hessian expression motivates the structured surrogate used in our update. In the implementation, we use the final approximation in Eq.~\ref{eq:hessian}, rather than the exact Hessian of the full learned objective:
\begin{equation}
\begin{aligned}
\mathbf{H}_t 
&= \nabla^{2}_{\mathbf{x}} J(\mathbf{x}_t) \\
&= \nabla_{\mathbf{x}}\left(\lambda_t \mathbf{A}^{T}(\mathbf{A}\mathbf{x}_t-\mathbf{y}) + \mathcal{G}_\theta(\mathbf{x}_t)\right) \\
&= \lambda_t \mathbf{A}^{T}\mathbf{A} + \nabla_{\mathbf{x}}\mathcal{G}_\theta(\mathbf{x}_t) \\
&\approx \lambda_t \mathbf{A}^{T}\mathbf{A} + \mathbf{I}.
\end{aligned}
\label{eq:hessian}
\end{equation}

This structured surrogate yields a symmetric positive definite system, which is particularly suitable for efficient optimization using the Conjugate Gradient (CG) method. It preserves the data-fidelity curvature and uses a stable identity approximation for the learned regularization term, rather than modeling its full curvature.

\subsection{Conjugate Gradient-Based Update}
\label{subsec:conjug}

\paragraph{Derivation of $b_t$.} To use CG for the structured surrogate update in Eq.~\ref{eq:newton}, we reformulate it as 
\begin{equation}
\mathbf{H}_t \mathbf{x}_{t+1} = b_t,
\end{equation}
where $b_t \in \mathbb{R}^{n}$ is the corresponding right-hand side vector.

Starting from the conceptual Newton-type form, we have
\begin{equation}
\mathbf{x}_{t+1} 
= \mathbf{x}_{t} - (\nabla^2_{\mathbf{x}} J(\mathbf{x}_t))^{-1}\nabla_{\mathbf{x}} J(\mathbf{x}_t).
\end{equation}

Multiplying both sides by $\mathbf{H}_t$ yields
\begin{equation}
\mathbf{H}_t \mathbf{x}_{t+1} 
= \mathbf{H}_t \mathbf{x}_t - \nabla_\mathbf{x} J(\mathbf{x}_t).
\end{equation}

Using the structured surrogate in Eq.~\ref{eq:hessian} and simplifying, we obtain
\begin{equation}
\mathbf{H}_t \mathbf{x}_{t+1}
= \lambda_t \mathbf{A}^{T} \mathbf{y} + \mathbf{x}_t - \mathcal{G}_\theta(\mathbf{x}_t)
= b_t.
\label{eq:second_opt}
\end{equation}

\noindent\textbf{Conjugate Gradient for the Structured Surrogate.} Directly computing the inverse of the surrogate operator is infeasible for high-dimensional CT images. Instead, we solve the linear system iteratively using the Conjugate Gradient (CG) method. CG is particularly suitable for large-scale problems because it only requires repeated applications of the linear operator $\mathbf{H}_t$ without explicitly forming the matrix. Starting from the current estimate $\mathbf{x}_t$, the CG solver generates a sequence of updates along mutually conjugate directions to approximate the solution of Eq.~\ref{eq:second_opt}. CG is used here as a linear-system solver; it does not estimate learned-prior curvature.
The procedure is summarized in Algorithm~\ref{alg:cg_update}, and the full unrolled CG-GLORE pipeline is illustrated in Figure~\ref{fig:overview}.

\begin{algorithm}[t]
\caption{Conjugate Gradient Reconstruction Update}
\label{alg:cg_update}

\textbf{Require:} current reconstruction $\mathbf{x}_t$, sinogram $\mathbf{y}$, projection operators $\mathbf{A}$ and $\mathbf{A}^T$, learned gradient module $\mathcal{G}_\theta(\cdot)$, number of CG iterations $K$

\begin{algorithmic}[1]

\State $b_t = \lambda_t \mathbf{A}^{T} \mathbf{y} + \mathbf{x}_t - \mathcal{G}_\theta(\mathbf{x}_t)$
\Comment{Construct right-hand side}

\State $\mathbf{x}^{(0)} = \mathbf{x}_t$
\Comment{Initialize reconstruction}

\State $r^{(0)} = b_t - \mathbf{H}_t \mathbf{x}^{(0)}$
\Comment{Compute residual}

\State $p^{(0)} = r^{(0)}$
\Comment{Initialize search direction}

\State $\rho^{0} = \|r^{(0)}\|^2$
\Comment{Compute residual norm squared}

\For{$k = 0,1,\dots,K-1$}
\State $w =
\mathbf{H}_t p^{k} =
\lambda_t \mathbf{A}^T\mathbf{A}p^{k} + p^{k}$
\Comment{Apply structured surrogate along search direction}

\State $\alpha^k =
\frac{\rho^k}{(p^{k})^\top w}$
\Comment{Compute step size}

\State $\mathbf{x}^{(k+1)} = \mathbf{x}^{k} + \alpha^k p^{k}$
\Comment{Update reconstruction}

\State $r^{(k+1)} = r^{k} - \alpha^k w$
\Comment{Update residual}

\State $\rho^{(k+1)} = \|r^{(k+1)}\|^2$
\Comment{Update residual norm squared}

\State $\beta^k = 
\frac{\rho^{(k+1)}}{\rho^{k}}$
\Comment{Compute conjugate coefficient}

\State $p^{(k+1)} = r^{(k+1)} + \beta^k p^{k}$
\Comment{Update search direction}

\EndFor

\State $\mathbf{x}_{t+1} = \mathbf{x}^{(K)}$
\Comment{Return updated reconstruction}

\end{algorithmic}
\end{algorithm}

\section{Experiments}

\subsection{Datasets \& Scenarios} 

\noindent\textbf{Datasets.} For evaluation, two datasets are used: ``2016 NIH-AAPM-Mayo Clinic Low-Dose CT Grand Challenge" (AAPM) \cite{aapm} and DeepLesion \cite{deeplesion}. The AAPM dataset includes $3mm$ thickness 2,378 full-dose CT images from 10 patients, while the DeepLesion dataset comprises CT images from 4,427 patients. For AAPM, we follow the strategy of \cite{ayad2024qn}, which selects images from 8 patients as the training set, 1 patient as the validation set, and 1 patient as the testing set. For DeepLesion, we use the official train-validation-test split, selecting 2,000, 200, and 300 CT images, respectively, for training and evaluation. All models are selected based on their best performance on the validation set, and results are reported on the testing set.

\subsection{Implementation Details} 

\noindent\textbf{Hyperparameters \& Libraries.} Given the original CT images, we resize them to $256 \times 256$ pixels, \textit{i.e.}, $H=256$ and $W=256$. We then use the ODL library \cite{adler2017operator} to simulate the forward process using a 2D fan-beam geometry with 512 detector pixels, a source-to-axis distance of 600 mm, and an axis-to-detector distance of 290 mm. The sinograms are generated over a full angular range of $[0, 2\pi]$ radians, resulting in 512 views. To simulate sparse-view scenarios, we select $n_{\text{view}} \in \{18, 32, 64, 128\}$ views, evenly spaced across the angular range. For reconstruction, we use the ODL backward process to apply the FBP algorithm. For the default setting, we set the sliding window size to $k=2$, which reduces the image resolution by a factor of $4$, resulting in $\tilde{N} = 128^2$. The number of landmarks is set as $L=384$. The three convolutional layers are configured with dimension sizes $C_1 = C_2 = C_3 = C = 64$. GLORE and all comparative models are trained for 50 epochs with a learning rate of $1 \times 10^{-4}$.

\noindent\textbf{Noise Configurations.} To simulate realistic scanning conditions, Poisson and Gaussian noise are added to the simulated sinograms. Poisson noise models photon-counting fluctuations, while Gaussian noise represents sensor and electronic noise. The noisy sinograms are regenerated at every training epoch to increase noise diversity and reduce overfitting. We evaluate three settings: No noise ($N_G=0, N_P=0$), Low noise ($N_G=5\%, N_P=10^6$), and High noise ($N_G=5\%, N_P=5\times10^5$), where lower $N_P$ corresponds to stronger Poisson noise.

\noindent\textbf{Noise Power Spectrum (NPS) evaluation.} The Noise Power Spectrum (NPS) characterizes the frequency distribution of noise in reconstructed CT images. For the $k$-th noise ROI $\mathbf{n}_k(x,y)$ of size $M\times N$, after mean subtraction and low-frequency trend removal, the 2D NPS is estimated as $\mathrm{NPS}(f_x,f_y)=\frac{\Delta_x\Delta_y}{KMN}\sum_{k=1}^{K}\left|\mathcal{F}{\mathbf{n}_k(x,y)}\right|^2$, where $\mathcal{F}{\cdot}$ denotes the 2D Fourier transform and $\Delta_x,\Delta_y$ are the pixel spacings. A rotationally invariant 1D NPS is obtained by radial averaging as $\mathrm{NPS}(f_r)=\frac{1}{|\Omega(f_r)|}\sum_{(f_x,f_y)\in\Omega(f_r)}\mathrm{NPS}(f_x,f_y)$, with $f_r=\sqrt{f_x^2+f_y^2}$. In our implementation, $M=N=32$, $K=57$, and $\Delta_x=\Delta_y=1.5625$ mm.

\subsection{Quantitative Results}

Following previous studies, we use Peak signal to noise ratio (PSNR) and structural index similarity (SSIM) to evaluate the quality of reconstructed CT image. 

\begin{table*}[h]
\resizebox{\textwidth}{!}{
\begin{tabular}{l|cccccc|cccccc|cccccc}
\toprule
\multirow{3}{*}{Method} 
& \multicolumn{6}{c|}{No noise}                                        
& \multicolumn{6}{c|}{$N_G$ = 5\%, $N_P$ = $10^6$}                           
& \multicolumn{6}{c}{$N_G$ = 5\%, $N_P$ = $5\times10^5$}                          \\ 
\cmidrule{2-19}

& \multicolumn{2}{c}{$n_{view}$ = 32} 
& \multicolumn{2}{c}{$n_{view}$ = 64} 
& \multicolumn{2}{c|}{$n_{view}$ = 128} 

& \multicolumn{2}{c}{$n_{view}$ = 32} 
& \multicolumn{2}{c}{$n_{view}$ = 64} 
& \multicolumn{2}{c|}{$n_{view}$ = 128} 

& \multicolumn{2}{c}{$n_{view}$ = 32} 
& \multicolumn{2}{c}{$n_{view}$ = 64} 
& \multicolumn{2}{c}{$n_{view}$ = 128} \\ 

\cmidrule{2-19}

& \textit{PSNR} & \textit{SSIM}  
& \textit{PSNR} & \textit{SSIM}  
& \textit{PSNR} & \textit{SSIM}

& \textit{PSNR} & \textit{SSIM}  
& \textit{PSNR} & \textit{SSIM}  
& \textit{PSNR} & \textit{SSIM}

& \textit{PSNR} & \textit{SSIM}  
& \textit{PSNR} & \textit{SSIM}  
& \textit{PSNR} & \textit{SSIM}  

\\ \midrule

FBPConvNet \cite{jin2017deep}             
& 31.33 & 87.02 & 38.42 & 94.02 & 42.97 & 97.46
& 31.52 & 83.37 & 37.35 & 94.32 & 42.06 & 97.34
& 31.28 & 86.6 & 36.81 & 93.57 & 41.23 & 96.83\\

LearnedPD \cite{adler2018learned}      
& \second{40.7} & \second{97.07} & 47.04 & 98.95 & 51.62 & 99.56
& 39.59 & 96.27 & 43.87 & 98.2 & 45.72 & 98.7
& 39.07 & 96.07 & 42.81 & 97.83 & 44.49 & 98.39\\

LEARN \cite{chen2018learn}          
& 40.52 & 96.78 & \second{47.65} & \second{99.04} & 50.52 & 99.47
& \second{40.04} & \second{96.56} & 44.09 & 98.24 & 45.82 & 98.71
& 39.53 & 96.25 & 42.98 & 97.88 & 44.58 & 98.4\\

QN-Mixer$^\dagger$ \cite{ayad2024qn}     
& 39.51 & 96.11 & 45.57 & 98.48 & 50.09 & 99.32
& 37.50 & 94.92 & 42.46 & 97.70 & 44.27 & 98.11
& 35.91 & 92.49 & 38.73 & 94.92 & 40.51 & 96.27\\

DPMA \cite{wu2025dual}     
& 35.2 & 92.16 & 40.11 & 96.63 & 45.74 & 98.89
& 34.82 & 90.64 & 38.96 & 94.93 & 42.6 & 97.23
& 34.7 & 90.05 & 38.04 & 93.29 & 41.03 & 95.75\\

RegFormer \cite{xia2023regformer}    
& 40.39 & 96.70 & 47.33 & 98.97 & \second{52.36} & \second{99.61}
& 39.62 & 96.28 & \best{44.24} & \best{98.29} & \best{46.35} & \best{98.83}
& 39.08 & 95.92 & \best{43.22} & \best{98.02} & \best{45.15} & \best{98.57}\\

\midrule
CG-GLORE (ours)          
& \best{42.05} & \best{97.59} & \best{47.66} & \best{99.05} & \best{52.43} & \best{99.62}
& \best{40.54} & \best{96.85} & \second{44.16} & \second{98.25} & \second{45.94} & \second{98.74}
& \best{39.86} & \best{96.47} & \second{43.07} & \second{97.93} & \second{44.76} & \second{98.46}\\

\bottomrule
\end{tabular}}
\footnotesize{$^\dagger$ The results are cited from \cite{ayad2024qn}.}
\vspace{-0.5em}
\caption{Quantitative comparison between \textcolor{black}{CG-GLORE} and state-of-the-art methods on the AAPM dataset. The best and second-best results are highlighted in red (bold) and yellow (underlined), respectively.}
\label{tab:aapm}
\end{table*}

\noindent\textbf{Results on AAPM dataset.}
Table \ref{tab:aapm} compares CG-GLORE with other methods on the AAPM dataset. In the noiseless setting, CG-GLORE achieves the best reconstruction quality across all projection-view numbers, reaching 42.05/97.59, 47.66/99.05, and 52.43/99.62 in PSNR/SSIM for 32, 64, and 128 views, respectively. The improvement is most pronounced in the highly sparse 32-view case, where CG-GLORE outperforms the second-best method by 1.53 dB in PSNR. Under noisy acquisition, CG-GLORE remains particularly robust at 32 views and obtains the best results under both noise levels. For 64 and 128 views, RegFormer slightly outperforms CG-GLORE in several noisy settings, but CG-GLORE consistently ranks second with close performance. These results indicate that the proposed CG-based update is especially beneficial in more ill-conditioned sparse-view regimes, while GLORE provides competitive regularization under different noise conditions.

\begin{table*}[t]
\centering
\footnotesize
\resizebox{\textwidth}{!}{
\begin{tabular}{l|cccc|cccc|cccc}
\toprule
\multirow{3}{*}{Method} 
& \multicolumn{4}{c|}{No noise}                                        
& \multicolumn{4}{c|}{$N_G$ = 5\%, $N_P$ = $10^6$}                           
& \multicolumn{4}{c}{$N_G$ = 5\%, $N_P$ = $5\times10^5$}                          \\ 
\cmidrule{2-13}

& \multicolumn{2}{c}{$n_{view}$ = 18} 
& \multicolumn{2}{c|}{$n_{view}$ = 32} 

& \multicolumn{2}{c}{$n_{view}$ = 18} 
& \multicolumn{2}{c|}{$n_{view}$ = 32} 

& \multicolumn{2}{c}{$n_{view}$ = 18} 
& \multicolumn{2}{c}{$n_{view}$ = 32} \\ 

\cmidrule{2-13}

& \textit{PSNR} & \textit{SSIM}  
& \textit{PSNR} & \textit{SSIM}

& \textit{PSNR} & \textit{SSIM}  
& \textit{PSNR} & \textit{SSIM}

& \textit{PSNR} & \textit{SSIM}  
& \textit{PSNR} & \textit{SSIM}  

\\ \midrule

FBPConvNet \cite{jin2017deep}             
& 28.72 & 69.2 & 32.76 & 84.89
& 29.27 & 79.77 & 32.21 & 82.27
& 29.1 & 78.53 & 32.74 & 84.46 \\

LearnedPD \cite{adler2018learned}      
& 34.82 & 92.34 & 39.90 & 96.50
& 34.65 & \second{92.23} & 39.37 & 96.14
& 34.54 & 92.04 & 39.09 & 96.00\\

DPMA \cite{wu2025dual}     
& 31.92 & 84.73 & 35.34 & 89.53
& 31.81 & 83.69 & 35.09 & 88.25
& 31.73 & 83.23 & 34.76 & 87.34\\

RegFormer \cite{xia2023regformer}    
& \second{35.64} & \second{92.49} & \second{40.71} & \second{96.81}
& \second{35.48} & 92.05 & \second{40.16} & \second{96.55}
& \second{35.39} & \second{92.19} & \second{39.80} & \second{96.33}\\

\midrule
CG-GLORE (ours)          
& \best{36.47} & \best{94.26} & \best{41.20} & \best{97.26}
& \best{36.35} & \best{94.21} & \best{40.52} & \best{96.96}
& \best{36.11} & \best{94.06} & \best{40.22} & \best{96.81}\\

\bottomrule
\end{tabular}}
\vspace{-0.5em}
\caption{Performance comparison between \textcolor{black}{CG-GLORE} and other state-of-the-art methods on DeepLesion dataset.}
\label{tab:deeplesion}
\end{table*}

\begin{figure}[t]
    \centering
    \includegraphics[width=1\textwidth]{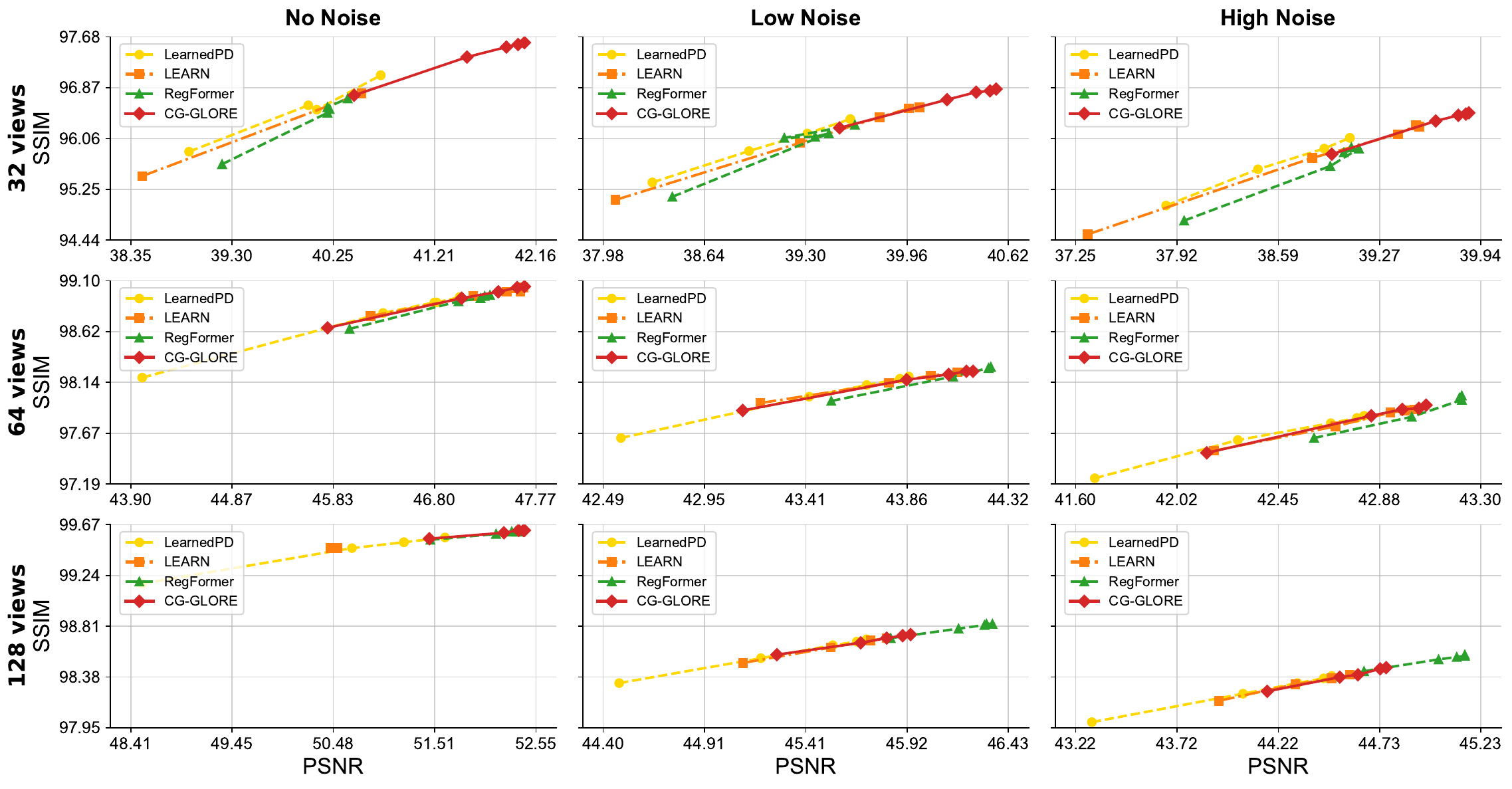}
    \vspace{-2em}
    \caption{\textbf{Convergence analysis on the AAPM dataset.} PSNR and SSIM are plotted as functions of training epochs for different reconstruction methods. Each curve consists of five evaluation points corresponding to 10, 20, 30, 40, and 50 training epochs, highlighting the convergence speed and stability of the compared methods.\vspace{-0.4cm}}
    \label{fig:convergence_aapm} 
\end{figure}

\noindent\textbf{Results on DeepLesion dataset.}
Table~\ref{tab:deeplesion} reports the performance on the DeepLesion dataset. CG-GLORE achieves the best PSNR and SSIM in all sparse-view and noise settings, demonstrating consistent robustness across both 18-view and 32-view reconstruction. In the challenging 18-view setting, CG-GLORE improves over the second-best method by 0.83 dB PSNR and 1.77 SSIM points without noise, and maintains clear margins under both noisy configurations. Similar gains are observed at 32 views, where CG-GLORE reaches 41.20/97.26 in the noiseless case and remains best under increasing Poisson noise. Together with the AAPM results, this consistent advantage suggests that the proposed global-local regularization and CG-based update provide robust reconstruction across different datasets and acquisition settings.

\noindent\textbf{Convergence analysis.}
Figure~\ref{fig:convergence_aapm} compares the training convergence of CG-GLORE with representative baselines on the AAPM dataset. CG-GLORE improves steadily from 10 to 50 epochs across all view and noise settings, showing stable optimization behavior. The clearest gains appear in the 32-view cases, where PSNR/SSIM increase from 40.45/96.75 to 42.05/97.59 without noise, from 39.52/96.23 to 40.54/96.85 under low noise, and from 38.95/95.81 to 39.86/96.47 under high noise. At 50 epochs, these results outperform the strongest plotted baseline by 1.35 dB, 0.50 dB, and 0.33 dB in PSNR, respectively. For 64 and 128 views, CG-GLORE remains best or competitive, especially in noise-free cases. The small gains from 40 to 50 epochs, at most 0.07 dB PSNR and 0.03 SSIM, further suggest stable late-stage convergence.

\begin{figure*}[t]
    \begin{minipage}[t]{0.49\textwidth}
        \centering
        \includegraphics[width=\linewidth]{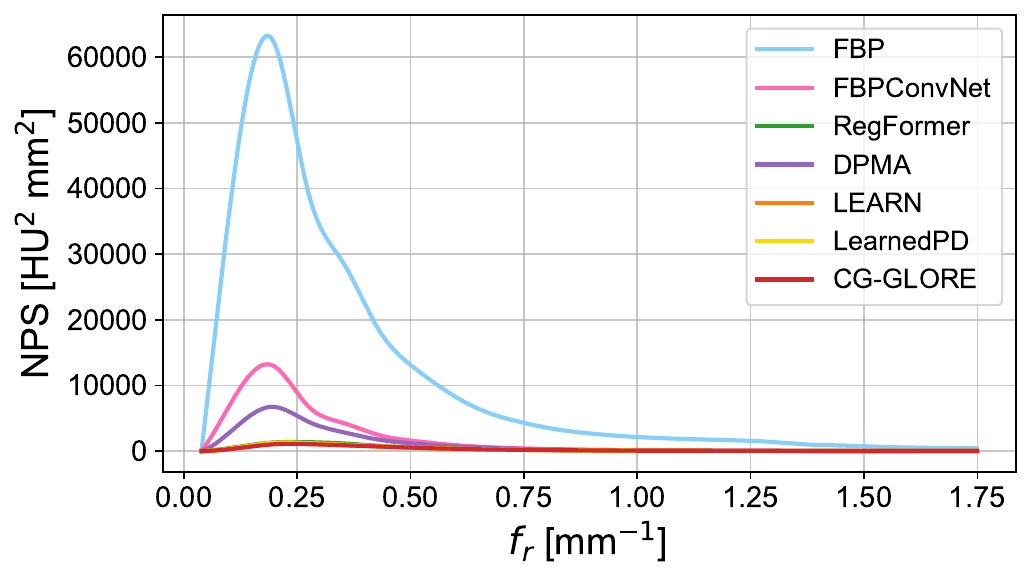}
        \vspace{-2.4em}
        \caption{NPS under low-noise setting \\ ($N_G=5\%$, $N_P=1\times10^6$) with $n_{view}=32$.}
        \label{fig:nps_32}
    \end{minipage}
    \begin{minipage}[t]{0.49\textwidth}
        \centering
        \includegraphics[width=\linewidth]{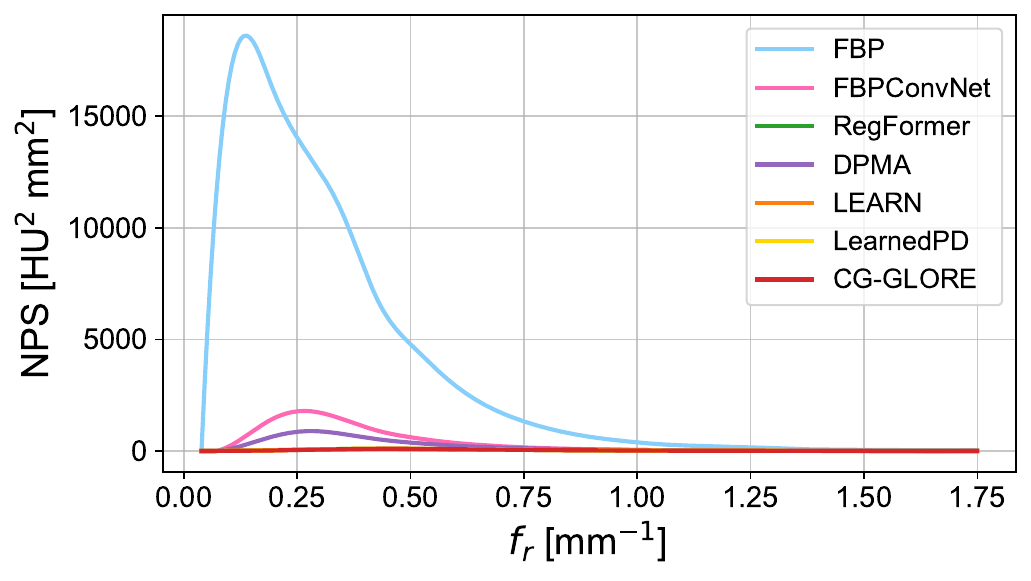}
        \vspace{-2.4em}
        \caption{NPS under noise-free setting \\($N_G=0$, $N_P=0$) with $n_{view}=64$.}
        \label{fig:nps_64}
    \end{minipage}

{\color{bmvcaptionblue}
\vspace{0.5em}
{\textbf{Figure \ref{fig:nps_32}-\ref{fig:nps_64}: ROIs for NPS Analysis.}
$32\times32$ ROIs are distributed as 1 center, 8 and 20 on two circular rings (radii 25, 50), and 28 on an outer ellipse (75×55), totaling 57 ROIs per image, within HU range $[-200, 300]$.}}
\end{figure*}
    
\noindent\textbf{Noise Power Spectrum analysis.}
Figure~\ref{fig:nps_32} and Figure~\ref{fig:nps_64} report the averaged 1D residual NPS curves computed from fixed ROIs over the test slices. In the more challenging 32-view setting, CG-GLORE achieves the lowest NPS over the full radial-frequency range, showing that it suppresses residual noise more effectively than the competing learned reconstruction methods. Under the 64-view setting, all advanced learning-based methods exhibit substantially reduced NPS compared with FBP and FBPConvNet, while CG-GLORE remains among the lowest curves and closely matches the best-performing alternatives. These results suggest that the CG-solved structured update improves reconstruction fidelity while reducing the frequency-dependent residual noise.

\begin{table*}[t]
\centering
\resizebox{0.7\linewidth}{!}{
\begin{tabular}{lccc}
\toprule
Method & \#Iters & Params (M) & Mem. (GB) \\
\midrule
FBPConvNet \cite{jin2017deep} & - & 31.1 & 1.51 \\
Learn PD \cite{adler2018learned} & 10 & 0.251 & 0.75 \\
LEARN \cite{chen2018learn} & 30 & 1.8 & 1.41 \\
QN-Mixer$^\dagger$ \cite{ayad2024qn} & 14 & 8.5 & 7.83 \\
RegFormer \cite{xia2023regformer} & 10 & 2.7 & 5.79 \\
DPMA \cite{wu2025dual} & 300 & 8.1 & 7.78 \\
\midrule
CG-GLORE (ours) & 15 & 5.5 & 5.6 \\
\bottomrule
\footnotesize{$^\dagger$ The results are cited from \cite{ayad2024qn}.}
\end{tabular}}
\vspace{1em}
\caption{Computational efficiency comparison of sparse-view CT reconstruction methods trained on AAPM with $n_{\text{view}}=32$.}
\label{tab:comparison}
\end{table*}

\noindent\textbf{Computational Efficiency Comparison.}
Table~\ref{tab:comparison} compares different reconstruction methods in terms of iteration count, model size, and memory usage when available. CG-GLORE uses 15 reconstruction iterations and 5.5M parameters, offering a practical balance between optimization strength and model complexity. Compared with LEARN and DPMA, CG-GLORE requires substantially fewer iterations (15 vs. 30 and 300), while its parameter count remains lower than FBPConvNet, QN-Mixer, and DPMA. Although lightweight methods such as Learned PD use fewer parameters, CG-GLORE achieves stronger reconstruction accuracy by using CG-solved structured updates together with the compact GLORE regularizer. These results suggest that CG-GLORE improves reconstruction quality without relying on excessive unrolling depth or a very large network.

\subsection{Qualitative Results}

\begin{figure}[t]
    \centering
    \includegraphics[width=1\textwidth]{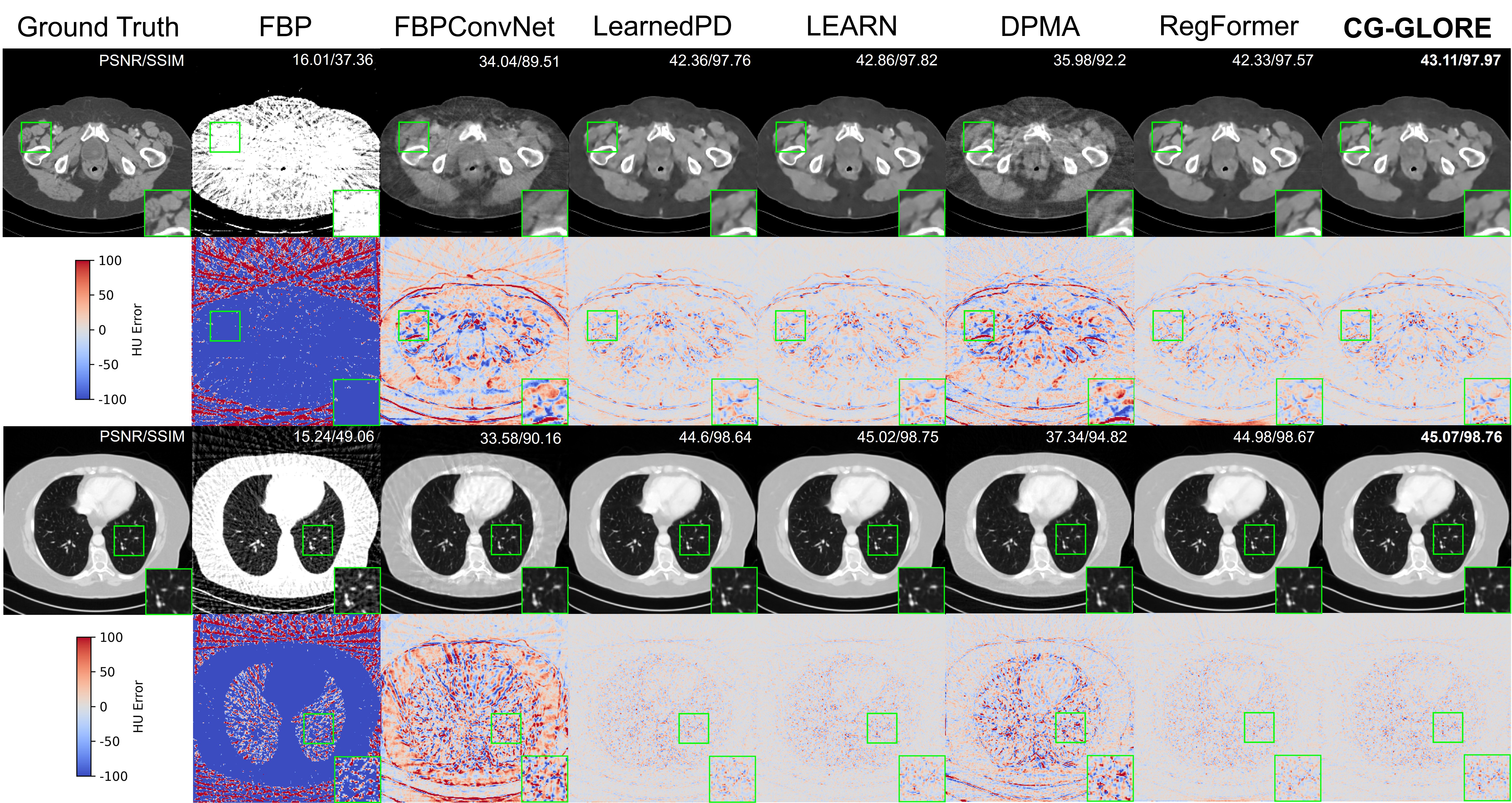}
    \vspace{-2em}
    \caption{\textbf{Visual comparison on the AAPM dataset.} The figure presents reconstructed CT images and their corresponding difference maps under different acquisition settings. \textbf{Top row:} sparse-view reconstruction with 32 projections and low noise level, visualized in the HU window $[-200, 300]$. \textbf{Bottom row:} reconstruction with 64 projections without noise, visualized in the HU window $[-1000, 700]$.}
    \label{fig:visual_aapm} 
\end{figure}

\noindent\textbf{Visualizations of Reconstructed CT Images.}
Figure~\ref{fig:visual_aapm} compares representative AAPM slices under two acquisition settings. FBP shows severe streak artifacts and large HU errors from limited-view measurements. Learning-based baselines reduce these artifacts, but several methods blur fine anatomy or retain structured residuals near high-contrast boundaries. In contrast, CG-GLORE produces reconstructions closer to the ground truth, with clearer soft-tissue structures in the 32-view low-noise case and sharper lung vessels in the 64-view noise-free case. The zoomed regions show better preservation of local edges and small structures, while the difference maps contain weaker residual patterns. These observations agree with the reported PSNR/SSIM values, where CG-GLORE achieves the best reconstruction quality in both examples. The visual results suggest that the CG-solved structured update improves data consistency, while GLORE recovers anatomical details without excessive smoothing.

\begin{figure}[t]
    \centering
    \includegraphics[width=1\textwidth]{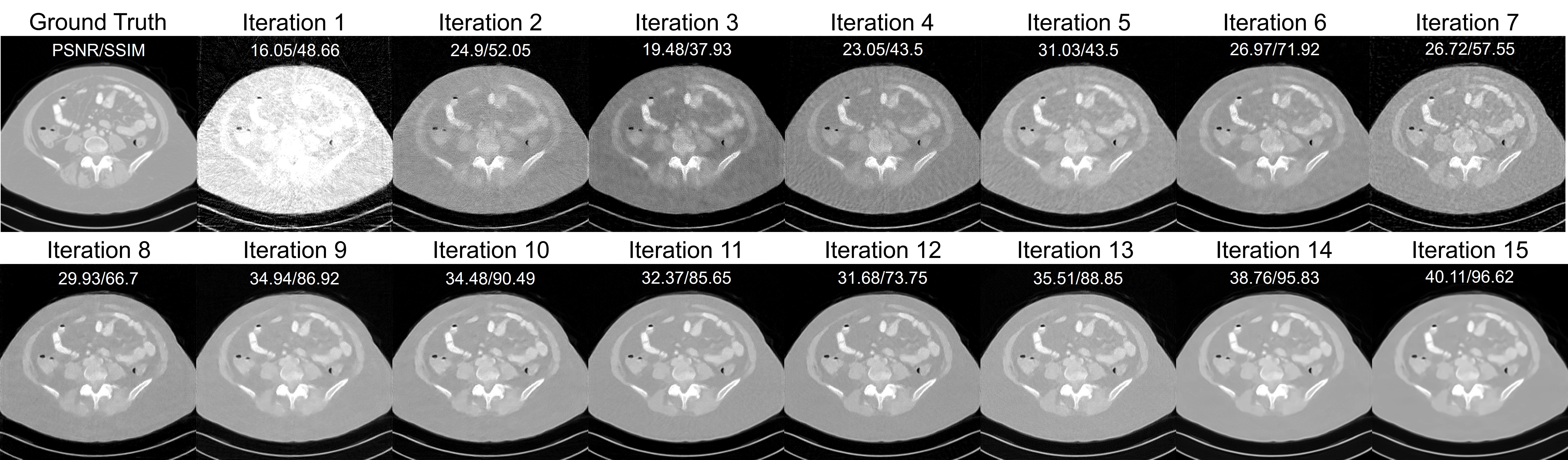}
    \vspace{-2em}
    \caption{\textbf{Stage-wise reconstruction evolution of CG-GLORE.} Intermediate reconstructions are shown under 32 projection views and a high-noise setting ($N_P=5 \times 10^5$), visualized in the HU window $[-1000, 700]$.\vspace{-0.6cm}}
    \label{fig:stagewise_reconstruction} 
\end{figure}

\noindent\textbf{Stage-wise Reconstruction Evolution.}
Figure~\ref{fig:stagewise_reconstruction} shows the intermediate outputs of CG-GLORE across the unrolled stages. Starting from degraded early reconstructions, the model progressively suppresses streak artifacts and recovers the global body contour, followed by clearer soft-tissue contrast and more stable anatomical boundaries. The PSNR/SSIM values generally increase with iteration, indicating that the CG solver and learned regularization jointly refine the reconstruction instead of producing a single abrupt correction. This stage-wise behavior supports the optimization interpretation of CG-GLORE: each unrolled step adds a data-consistent refinement under the learned global-local prior.

\subsection{Ablation Studies}

\begin{table*}[t]
\label{tab:ablation_studies}

\begin{minipage}[t]{0.4\textwidth}
\centering
\resizebox{\linewidth}{!}{
\begin{tabular}{lcc}
\toprule
Variant & PSNR & SSIM \\
\midrule
3 Convs + First-order & 36.02 & 92.94 \\
GLORE + First-order & 40.09 & 96.63 \\
3 Convs + CG  & 40.45 & 96.76 \\
GLORE + CG & \textbf{40.54} & \textbf{96.85} \\
\bottomrule
\end{tabular}}
\vspace{-0.9em}
\caption{Impact of CG and GLORE components.}
\label{tab:ablation_design}
\end{minipage}
\begin{minipage}[t]{0.275\textwidth}
\centering
\resizebox{\linewidth}{!}{
\begin{tabular}{ccc}
\toprule
CG Iter. & PSNR & SSIM \\
\midrule
1 & 33.95 & 83.29 \\
3 & 40.29 & 96.72 \\
5 & 40.45 & 96.78 \\
10 & \textbf{40.54} & \textbf{96.85} \\
\bottomrule
\end{tabular}}
\vspace{-0.9em}
\caption{Impact of CG solver depth.}
\label{tab:ablation_cg_iter}
\end{minipage}
\begin{minipage}[t]{0.3\textwidth}
\centering
\resizebox{\linewidth}{!}{
\begin{tabular}{ccc}
\toprule
$(k \times k, C)$ & PSNR & SSIM \\
\midrule
$(1 \times 1, 64)$ & 40.45 & 96.79 \\
$(2 \times 2, 64)$ & \textbf{40.54} & \textbf{96.85} \\
$(4 \times 4, 32)$ & 39.22 & 96.07 \\
$(8 \times 8, 24)$ & 37.21 & 94.59 \\
\bottomrule
\end{tabular}}
\vspace{-0.9em}
\caption{Impact of Sliding Window $(k \times k)$ and Embedding Size $C$.}
\label{tab:ablation_patch}
\end{minipage}
{\color{bmvcaptionblue}
\vspace{0.4em}
\noindent\textbf{Tables~\ref{tab:ablation_design}--\ref{tab:ablation_patch}: Component and hyperparameter analysis of CG-GLORE on AAPM} under the shared setting of $n_{view}=32$, $N_G=5\%$, and $N_P=1\times 10^6$.\par}
\end{table*}

\noindent\textbf{CG-GLORE Ablation.}
Table~\ref{tab:ablation_design} evaluates the contribution of the CG-based update and the GLORE regularizer. Using only three convolutional layers with a first-order update gives the weakest result, indicating that a shallow local prior alone is insufficient for sparse-view reconstruction. Replacing the local regularizer with GLORE under the same first-order update improves PSNR from 36.02 to 40.09 and SSIM from 92.94 to 96.63, showing the benefit of combining local features with long-range dependency modeling. Introducing CG with the three-convolution regularizer also yields a large improvement, reaching 40.45/96.76 in PSNR/SSIM. The full CG-GLORE model achieves the best result of 40.54/96.85, confirming that the CG-solved structured update and global-local regularization are complementary.

\noindent\textbf{CG Solver Depth.}
Table~\ref{tab:ablation_cg_iter} studies the number of CG iterations used to solve the structured surrogate system. With only one CG iteration, the approximation is insufficient and the reconstruction quality remains low. Increasing the depth to three iterations produces a substantial gain, from 33.95/83.29 to 40.29/96.72, indicating that a few conjugate directions already provide a more accurate surrogate update. Further increasing the depth to five and ten iterations gives smaller but consistent improvements. We therefore use ten CG iterations as the default setting, which offers the best reconstruction quality while keeping the solver cost moderate.

\noindent\textbf{Sliding Window Size and Embedding Dimension.}
Table~\ref{tab:ablation_patch} analyzes the sparse patchification design in LORAD. The $1 \times 1$ setting preserves the full spatial resolution but provides limited token-level aggregation. The proposed $2 \times 2$ window achieves the best performance, suggesting a favorable balance between long-range context modeling and local detail preservation. Larger windows, such as $4 \times 4$ and $8 \times 8$, reduce the number of tokens more aggressively and require smaller embedding dimensions, which weakens the representation of fine structures and leads to degraded reconstruction quality. This supports the choice of $k=2$ and $C=64$ in the default CG-GLORE configuration.

\section{Conclusion}
We presented \textbf{CG-GLORE}, a second-order-inspired deep unrolling framework for sparse-view CT reconstruction. Each stage uses a CG-solvable structured Hessian surrogate that retains the physics-induced data-fidelity curvature while using an identity approximation for the learned regularization term; it is therefore not an exact Newton method for the full learned objective. GLORE combines convolutional local modeling with efficient long-range dependency representation. Experiments on AAPM and DeepLesion show improved reconstruction quality, particularly in highly ill-conditioned settings, while ablations support the complementary contributions of the CG solver and GLORE. Overall, CG-GLORE offers an effective balance between model-based optimization and learned regularization for sparse-view CT reconstruction. Future work will investigate broader validation on real clinical acquisition protocols and extension to more general 3D reconstruction settings.

\section*{Acknowledgment}
This work was supported by JST-ALCA-Next (JPMJAN23F4).

\bibliography{egbib}

@article{wang2008outlook,
  title={An outlook on x-ray CT research and development},
  author={Wang, Ge and Yu, Hengyong and De Man, Bruno},
  journal={Medical physics},
  volume={35},
  number={3},
  pages={1051--1064},
  year={2008},
  publisher={Wiley Online Library}
}

@article{katsura2012model,
  title={Model-based iterative reconstruction technique for radiation dose reduction in chest CT: comparison with the adaptive statistical iterative reconstruction technique},
  author={Katsura, Masaki and Matsuda, Izuru and Akahane, Masaaki and Sato, Jiro and Akai, Hiroyuki and Yasaka, Koichiro and Kunimatsu, Akira and Ohtomo, Kuni},
  journal={European radiology},
  volume={22},
  number={8},
  pages={1613--1623},
  year={2012},
  publisher={Springer}
}

@article{cormack1963representation,
  title={Representation of a function by its line integrals, with some radiological applications},
  author={Cormack, Allan Macleod},
  journal={Journal of applied physics},
  volume={34},
  number={9},
  pages={2722--2727},
  year={1963},
  publisher={American Institute of Physics}
}

@article{chao2022sparse,
  title={Sparse-view cone beam CT reconstruction using dual CNNs in projection domain and image domain},
  author={Chao, Lianying and Wang, Zhiwei and Zhang, Haobo and Xu, Wenting and Zhang, Peng and Li, Qiang},
  journal={Neurocomputing},
  volume={493},
  pages={536--547},
  year={2022},
  publisher={Elsevier}
}

@article{wang2024low,
  title={Low-dose CT reconstruction using dataset-free learning},
  author={Wang, Feng and Wang, Renfang and Qiu, Hong},
  journal={PLoS One},
  volume={19},
  number={6},
  pages={e0304738},
  year={2024},
  publisher={Public Library of Science San Francisco, CA USA}
}

@article{tao2021learning,
  title={Learning to reconstruct CT images from the VVBP-tensor},
  author={Tao, Xi and Wang, Yongbo and Lin, Liyan and Hong, Zixuan and Ma, Jianhua},
  journal={IEEE Transactions on Medical Imaging},
  volume={40},
  number={11},
  pages={3030--3041},
  year={2021},
  publisher={IEEE}
}

@article{galantai2000theory,
  title={The theory of Newton's method},
  author={Gal{\'a}ntai, Aurel},
  journal={Journal of Computational and Applied Mathematics},
  volume={124},
  number={1-2},
  pages={25--44},
  year={2000},
  publisher={Elsevier}
}

@article{wu2021noise,
author = {Wu, Dufan and Kim, Kyungsang and Li, Quanzheng},
title = {Low-dose CT reconstruction with Noise2Noise network and testing-time fine-tuning},
journal = {Medical Physics},
volume = {48},
number = {12},
pages = {7657-7672},
year = {2021}
}

@article{jin2017deep,
  title={Deep convolutional neural network for inverse problems in imaging},
  author={Jin, Kyong Hwan and McCann, Michael T and Froustey, Emmanuel and Unser, Michael},
  journal={IEEE transactions on image processing},
  volume={26},
  number={9},
  pages={4509--4522},
  year={2017},
  publisher={IEEE}
}

@article{adler2018learned,
  title={Learned primal-dual reconstruction},
  author={Adler, Jonas and {\"O}ktem, Ozan},
  journal={IEEE transactions on medical imaging},
  volume={37},
  number={6},
  pages={1322--1332},
  year={2018},
  publisher={IEEE}
}

@article{chen2018learn,
  title={LEARN: Learned experts’ assessment-based reconstruction network for sparse-data CT},
  author={Chen, Hu and Zhang, Yi and Chen, Yunjin and Zhang, Junfeng and Zhang, Weihua and Sun, Huaiqiang and Lv, Yang and Liao, Peixi and Zhou, Jiliu and Wang, Ge},
  journal={IEEE transactions on medical imaging},
  volume={37},
  number={6},
  pages={1333--1347},
  year={2018},
  publisher={IEEE}
}

@inproceedings{ayad2024qn,
  title={QN-Mixer: A Quasi-Newton MLP-Mixer Model for Sparse-View CT Reconstruction},
  author={Ayad, Ishak and Larue, Nicolas and Nguyen, Ma{\"\i} K},
  booktitle={Proceedings of the IEEE/CVF Conference on Computer Vision and Pattern Recognition},
  pages={25317--25326},
  year={2024}
}

@article{xia2023regformer,
  title={RegFormer: A Local--Nonlocal Regularization-Based Model for Sparse-View CT Reconstruction},
  author={Xia, Wenjun and Yang, Ziyuan and Lu, Zexin and Wang, Zhongxian and Zhang, Yi},
  journal={IEEE Transactions on Radiation and Plasma Medical Sciences},
  volume={8},
  number={2},
  pages={184--194},
  year={2023},
  publisher={IEEE}
}

@article{wu2025dual,
  title={Dual-Domain deep prior guided sparse-view CT reconstruction with multi-scale fusion attention},
  author={Wu, Jia and Lin, Jinzhao and Jiang, Xiaoming and Zheng, Wei and Zhong, Lisha and Pang, Yu and Meng, Hongying and Li, Zhangyong},
  journal={Scientific Reports},
  volume={15},
  number={1},
  pages={16894},
  year={2025},
  publisher={Nature Publishing Group UK London}
}

@article{nguyen2024image,
  title={An image is worth more than 16x16 patches: Exploring transformers on individual pixels},
  author={Nguyen, Duy-Kien and Assran, Mahmoud and Jain, Unnat and Oswald, Martin R and Snoek, Cees GM and Chen, Xinlei},
  journal={arXiv preprint arXiv:2406.09415},
  year={2024}
}

@article{vaswani2017attention,
  title={Attention is all you need},
  author={Vaswani, Ashish and Shazeer, Noam and Parmar, Niki and Uszkoreit, Jakob and Jones, Llion and Gomez, Aidan N and Kaiser, {\L}ukasz and Polosukhin, Illia},
  journal={Advances in neural information processing systems},
  volume={30},
  year={2017}
}

@article{wang2013improving,
  title={Improving CUR matrix decomposition and the Nystr{\"o}m approximation via adaptive sampling},
  author={Wang, Shusen and Zhang, Zhihua},
  journal={The Journal of Machine Learning Research},
  volume={14},
  number={1},
  pages={2729--2769},
  year={2013},
  publisher={JMLR. org}
}

@inproceedings{xiong2021nystromformer,
  title={Nystr{\"o}mformer: A nystr{\"o}m-based algorithm for approximating self-attention},
  author={Xiong, Yunyang and Zeng, Zhanpeng and Chakraborty, Rudrasis and Tan, Mingxing and Fung, Glenn and Li, Yin and Singh, Vikas},
  booktitle={Proceedings of the AAAI conference on artificial intelligence},
  volume={35},
  number={16},
  pages={14138--14148},
  year={2021}
}

@article{shen2018baseline,
  title={Baseline needs more love: On simple word-embedding-based models and associated pooling mechanisms},
  author={Shen, Dinghan and Wang, Guoyin and Wang, Wenlin and Min, Martin Renqiang and Su, Qinliang and Zhang, Yizhe and Li, Chunyuan and Henao, Ricardo and Carin, Lawrence},
  journal={arXiv preprint arXiv:1805.09843},
  year={2018}
}

@inproceedings{razavi2014new,
  title={A new iterative method for finding approximate inverses of complex matrices},
  author={Razavi, M Kafaei and Kerayechian, Asghar and Gachpazan, Mortaza and Shateyi, Stanford},
  booktitle={Abstract and Applied Analysis},
  volume={2014},
  number={1},
  pages={563787},
  year={2014},
  organization={Wiley Online Library}
}

@article{aapm,
  title={TU-FG-207A-04: overview of the low dose CT grand challenge},
  author={McCollough, Cynthia},
  journal={Medical physics},
  volume={43},
  number={6Part35},
  pages={3759--3760},
  year={2016},
  publisher={Wiley Online Library}
}

@article{deeplesion,
  title={DeepLesion: automated mining of large-scale lesion annotations and universal lesion detection with deep learning},
  author={Yan, Ke and Wang, Xiaosong and Lu, Le and Summers, Ronald M},
  journal={Journal of medical imaging},
  volume={5},
  number={3},
  pages={036501--036501},
  year={2018},
  publisher={Society of Photo-Optical Instrumentation Engineers}
}

@article{adler2017operator,
  title={Operator discretization library (ODL)},
  author={Adler, Jonas and Kohr, Holger and {\"O}ktem, Ozan},
  journal={Zenodo},
  year={2017}
}

@inproceedings{he2016identity,
  title={Identity mappings in deep residual networks},
  author={He, Kaiming and Zhang, Xiangyu and Ren, Shaoqing and Sun, Jian},
  booktitle={ECCV},
  year={2016}
}

@article{levenberg1944method,
  title={A method for the solution of certain non-linear problems in least squares},
  author={Levenberg, Kenneth},
  journal={Quarterly of Applied Mathematics},
  year={1944}
}

@article{marquardt1963algorithm,
  title={An algorithm for least-squares estimation of nonlinear parameters},
  author={Marquardt, Donald},
  journal={SIAM Journal},
  year={1963}
}

@inproceedings{wang2022dudotrans,
  title={DuDoTrans: dual-domain transformer for sparse-view CT reconstruction},
  author={Wang, Ce and Shang, Kun and Zhang, Haimiao and Li, Qian and Zhou, S Kevin},
  booktitle={International Workshop on Machine Learning for Medical Image Reconstruction},
  pages={84--94},
  year={2022},
  organization={Springer}
}

@article{su2022generalized,
  title={Generalized deep iterative reconstruction for sparse-view CT imaging},
  author={Su, Ting and Cui, Zhuoxu and Yang, Jiecheng and Zhang, Yunxin and Liu, Jian and Zhu, Jiongtao and Gao, Xiang and Fang, Shibo and Zheng, Hairong and Ge, Yongshuai and others},
  journal={Physics in Medicine \& Biology},
  volume={67},
  number={2},
  pages={025005},
  year={2022},
  publisher={IOP Publishing}
}

@book{nocedal2006numerical,
  title={Numerical Optimization},
  author={Nocedal, Jorge and Wright, Stephen J.},
  edition={2},
  year={2006},
  publisher={Springer}
}

@article{royer2020newtoncg,
  title={A Newton-CG algorithm with complexity guarantees for smooth unconstrained optimization},
  author={Royer, Cl{\'e}ment W. and O'Neill, Michael and Wright, Stephen J.},
  journal={Mathematical Programming},
  volume={180},
  pages={451--488},
  year={2020}
}

@article{hestenes1952cg,
  title={Methods of conjugate gradients for solving linear systems},
  author={Hestenes, Magnus R. and Stiefel, Eduard},
  journal={Journal of Research of the National Bureau of Standards},
  volume={49},
  number={6},
  pages={409--436},
  year={1952}
}

\end{document}